# Fuzzy Rule-based Differentiable Representation Learning

Wei Zhang, Zhaohong Deng, *Senior Member, IEEE*, Guanjin Wang *Member, IEEE*, Kup-Sze Choi, *Senior Member*

*Abstract*—Representation learning has emerged as a crucial focus in machine and deep learning, involving the extraction of meaningful and useful features and patterns from the input data, thereby enhancing the performance of various downstream tasks such as classification, clustering, and prediction. Current mainstream representation learning methods primarily rely on non-linear data mining techniques such as kernel methods and deep neural networks (DNNs) to extract abstract knowledge from complex datasets. However, most of these methods are black-box, lacking transparency and interpretability in the learning process, which constrains their practical utility. To this end, this paper introduces a novel representation learning method grounded in an interpretable fuzzy rule-based model. Specifically, it is built upon the Takagi-Sugeno-Kang fuzzy system (TSK-FS) to initially map input data to a high-dimensional fuzzy feature space through the antecedent part of the TSK-FS. Subsequently, a novel differentiable optimization method is proposed for the consequence part learning which can preserve the model's interpretability and transparency while further exploring the nonlinear relationships within the data. This optimization method retains the essence of traditional optimization, with certain parts of the process parameterized corresponding differentiable modules constructed and a deep optimization process implemented. Consequently, this method not only enhances the model's performance but also ensures its interpretability. Moreover, a second-order geometry preservation method is introduced to further improve the robustness of the proposed method. Extensive experiments conducted on various benchmark datasets validate the superiority of the proposed method, highlighting its potential for advancing representation learning methodologies.

*Index Term*s – rule-based model, representation learning, fuzzy system, differentiable optimization.

## I. Introduction

In the last two decades, machine learning has undergone significant development, with a multitude of researchers delving into cutting-edge modeling structures to improve their robustness across various learning tasks [1, 2]. Notably, one critical aspect underpinning the performance of machine learning models is the quality of the data [3]. By acquiring efficient data representations, machines can extract meaningful and relevant features and patterns from the original input data, thereby facilitating improved decision-making capabilities in the downstream tasks. As a result, research on representation learning has garnered considerable attention in the field of machine and deep learning, ranging from linear transformation-based methods like Principal Component Analysis (PCA) [4], and Non-negative Matrix Factorization (NMF) [5] to nonlinear transformation-based methods like deep neural networks [6], and deep autoencoder [7].

With the increasing complexity of collected data and the advancement in computational capabilities in recent years, nonlinear transformation-based representation learning methods have progressively taken center stage. For example, Schölkopf et al. proposed Kernel Principal Component Analysis (KPCA) based on the kernel method, facilitating the mapping of data into a high-dimensional kernel space and subsequent extraction of a low-dimensional representation [8]. Similarly, methods like Kernel non-negative matrix factorization (KNMF) [9] combine the kernel method with NMF, which first maps data to kernel space and then factorizes it into a non-negative representation. Deep neural network (DNN)-based representation learning methods have also garnered attention for their ability to extract hierarchical discriminative information from complex data and condense it into a low-dimensional representation that retain the most relevant discriminative information [6]. Another example is the deep autoencoder (DAE) [7], which comprises encoder and decoder architectures constructed with DNNs to extract discriminative information. Moreover, several improved models including Variational Autoencoders (VAEs) [10], have been proposed to augment the performance of DAEs. On the other hand, in the realm of graph data, Li et al. introduced a graph convolutional network for analyzing such data structures [11], while the introduction of the attention mechanism led to the graph network [12]. Although these nonlinear transformation-based representation learning methods have achieved great success, their performance are constrained by their inherent 'black-box' nature, stemming from the lack of interpretability and transparency in kernel methods and deep neural networks. This limitation poses challenges for their application in scenarios that demand high levels of transparency and interpretability, such as medicine, finance and politics.

Rule-based modeling is one of the proven options for achieving the requirements of high interpretability and

This work was supported in part by the National key R & D plan under Grant (2022YFE0112400), the NSFC under Grant 62176105, the Six Talent Peaks Project in Jiangsu Province under Grant XYDXX-056). (Corresponding author: Zhaohong Deng).

W. Zhang is with the School of Artificial Intelligence and Computer Science, Nantong University, Nantong 226019, China and the School of Artificial Intelligence and Computer Science, Jiangnan University Wuxi 214122, China (e-mail: weizhang@ntu.edu.cn).

Z. Deng is with the School of Artificial Intelligence and Computer Science, Jiangnan University and Jiangsu Key Laboratory of Media Design and Software Technology, Wuxi 214122, China (e-mail: dengzhaohong@jiangnan.edu.cn).

G. Wang is with the School of Information Technology, Murdoch University, WA, Australia (Guanjin.Wang@murdoch.edu.au).

K. S. Choi is with The Centre for Smart Health, the Hong Kong Polytechnic University, Hong Kong (e-mail: kschoi@ieee.org).

transparency [13]. In particular, the Takagi-Sugeno-Kang fuzzy system (TSK-FS) is one of the rule models that has received lots of attention due to its strong data-driven ability [14, 15]. Various TSK-FS-based representation learning methods have been proposed and applied across different domains [16-18]. These methods can be summarized by following a process that first involves a nonlinear mapping using the antecedent parameters of TSK-FS, followed by the learning of linear consequence parameters to reduce the feature dimensions and extract knowledge. For example, in the context of transfer learning, Xu et al. used the antecedent parameters of TSK-FS to map the source and target domain data to a high-dimensional space followed by the learning of linear consequence parameters for source and target domain respectively, and performed feature alignment in low-dimensional spaces to realize feature transfer learning [16]. Similarly, Li et al. extended the method in [16] to include multi-source transfer learning and introduced multi-view distribution matching and structure preservation mechanisms to improve performance [17]. In addition, based on TSK-FS, Zhang et al. proposed a multi-view representation learning method, which mapped multi-view data to a high-dimensional space by using the antecedent parameters first, and then learned two parts of linear consequence parameters to explore the common and specific information in the multi-view data respectively [18].

These TSK-FS-based methods emerged as a promising research direction in constructing a reliable and interpretable representation learning. However, it is important to note that these methods are limited to specific domains, such as transfer learning or multi-view learning, highlighting the need to establish a more generalized representation learning method. In addition, the effectiveness of existing TSK-FS based methods for representation learning mainly depends on the antecedent parameters. If these parameters fail to map the original data into a discriminative space, it is still difficult to extract a high-quality low-dimensional representation during the learning process of the linear consequence parameters. To address this, recent studies have integrated DNNs as replacements for linear consequence parameters to improve model performance [19-21]. However, the direct use of DNNs as a part of the model reduces the transparency and interpretability of the learning process to some extent, due to the black-box nature of adopted DNNs. Therefore, constructing a representation learning method that ensures both superior performance and interpretability remain a significant research challenge.

To address the challenges highlighted above, this paper introduces a novel rule-based differentiable representation learning method based on TSK-FS. Taking the traditional TSK-FS as a basic model in the proposed method, the initial phase involves the construction of the antecedent part of the TSK-FS to facilitate nonlinear transformation and the mapping of data in the original space into a high-dimension fuzzy space. Then, a new deep consequent parameters learning method is proposed to achieve dimensionality reduction and obtain a higher discriminative low-dimensional representation. The proposed method preserves the traditional optimization process of consequent parameters, yet it introduces parameterized elements and develops corresponding differentiable modules within the process. Subsequently, a deep optimization procedure is implemented for consequent parameters. This method effectively establishes a linkage between the traditional TSK-FS and DNN, improves the nonlinear data mining ability while ensuring transparency and interpretability in the constructed rule model. Specifically, both the antecedent and consequence parts of TSK-FS demonstrate nonlinear mining ability, while retaining the traditional optimization process. Moreover, a second-order Laplacian graph method is incorporated which serves to preserve the geometry structure of the data, and further contributes to the improved discriminability of the learned low-dimensional representation.

The contributions of this paper include the following:
1) A novel fuzzy rule-based differentiable representation learning method is proposed, which not only expands the scope of application for traditional TSK-FS but also introduces a fresh perspective on connecting TSK-FS and DNN.
2) A new differentiable optimization method for learning consequence parameters is proposed, which can enhance the ability to extract patterns from the nonlinear complex data while preserving the transparency and interpretability of the proposed method.
3) The effectiveness of the proposed method is validated on comprehensive experimental studies.

The remainder of this paper is organized as follows. Section II provides a brief overview of the background. The proposed method is described in detail in Section III. Experimental studies and results are reported in Section IV. Finally, conclusions are drawn in Section V.

## II. BACKGROUND

In this section, we begin by introducing the traditional TKS-FS and deep TKS-FS models, following this, we delve into the iterative shrinkage-thresholding differentiable optimization method.

### A. Takagi-Sugeno-Kang Fuzzy System

As a rule-based intelligent model, TSK-FS has achieved great attention in the past decades due to its strong data-driven ability and high interpretability [22].

In the TSK-FS, the $h$th fuzzy rule is defined as follows:

$$IF\ x_1\ is\ A_1^h\ \wedge \cdots \wedge\ x_d\ is\ A_d^h$$
$$THEN\ f^h(\mathbf{x}) = p_0^h + p_1^h x_1 + \cdots + p_d^h x_d \quad (1)$$

where $h = 1,2,\ldots,H$, $H$ is the number of rules, $\mathbf{x} = [x_1, x_2, \ldots, x_d] \in R^{1\times d}$ is the input vector, $d$ is the number of features of the input, $f^k(\mathbf{x})$ is the output of the $h$th rule, $\wedge$ is a fuzzy conjunction operator, and $A_j^h$ is a fuzzy set associated with the $j$th feature and the $h$th rule. Usually, the choice of membership function of the fuzzy set in TSK-FS is important, and it can be defined as the Gaussian function when the specific domain knowledge is not available. In this scene, the membership value of each feature of the corresponding fuzzy set $A_j^h$ can be evaluated with the following:

$$\mu_{A_j^h}(x_j) = exp\left(-(x_j - e_{h,j})^2 / 2q_{h,j}\right) \tag{2a}$$

where parameters $e_{h,j}$ and $q_{h,j}$ are the center and width of the Gaussian function, respectively. These two parameters are also defined as the antecedent parameters of TSK-FS, and they can be evaluated by clustering algorithms, such as FCM [23]. Then, the firing level of the *h*th rule for the input vector can be obtained by concatenating the membership value of each feature, and it is shown in (2b). Meanwhile, the normalized form of (2b) is shown in (2c).

$$\mu^h(\mathbf{x}) = \prod_{j=1}^d \mu_{A_j^h}(x_j) \tag{2b}$$

$$\tilde{\mu}^h(\mathbf{x}) = \frac{\mu^h(\mathbf{x})}{\sum_{h'=1}^h \mu^{h'}(\mathbf{x})} \tag{2c}$$

Based on the above, the output of TSK-FS is as follows:

$$y = \sum_{h=1}^H \tilde{\mu}^h(\mathbf{x}) f^h(\mathbf{x}) \tag{2d}$$

However, (2d) is still difficult to optimize. Fortunately, it can be transformed into the form of linear regression in a new fuzzy feature space as follows:

$$y = \mathbf{x}_g \mathbf{p}_g \tag{3a}$$

where $\mathbf{x}_g$ and $\mathbf{p}_g$ are defined as follows:

$$\mathbf{x}_e = [1, \mathbf{x}] \in R^{1 \times (d+1)} \tag{3b}$$

$$\tilde{\mathbf{x}}^h = \tilde{\mu}^h(\mathbf{x})\mathbf{x}_e \in R^{1 \times (d+1)} \tag{3c}$$

$$\mathbf{x}_g = [\tilde{\mathbf{x}}^1, \tilde{\mathbf{x}}^2, ..., \tilde{\mathbf{x}}^H] \in R^{1 \times H(d+1)} \tag{3d}$$

$$\mathbf{p}_h = [p_0^h, p_1^h, ..., p_d^h] \in R^{1 \times (d+1)} \tag{3e}$$

$$\mathbf{p}_g = [\mathbf{p}_1, \mathbf{p}_2, ..., \mathbf{p}_H]^T \in R^{H(d+1) \times 1} \tag{3f}$$

*B. Deep Takagi-Sugeno-Kang Fuzzy System*

In recent years, there is a surge of interest in enhancing the performance of the TSK-FS model through constructing the deep learning frameworks, and leading to the development of several deep ensemble TSK-FSs. For example, Zhou et al. proposed a deep TSK-FS (D-TSK-FS), which consists of a few TSK-FS in a stacking way [24]. Building upon the D-TSK-FS framework, several new stacking-based deep TSK-FS models have been proposed to address the challenges in imbalanced data classification [25, 26]. Furthermore, an adversarial deep TSK-FS model was proposed, by incorporating smooth gradient information from the adversarial outputs of each layer [27]. In addition, to recognize the sleep-wake stages of persons, Zhou et al. proposed a random rule heritage based deep TSK-FS [28].

In addition, researchers have explored the integration of TSK-FS and DNNs in both sequential and parallel configurations to realize deep fuzzy system modelling. For example, Tian et al. proposed a deep TSK-FS by sequentially integrating TSK-FS and DNNs [29]. This approach first employs DNN to extract the features from EEG data, followed by inputting these features into the TSK-FS for classification purposes. Similarly, Li et al. proposed a deep transfer model with TSK-FS, wherein the DNN serves as the feature extraction component, and the TSK-FS is utilized as the classification model [30]. In addition, Deng et al. proposed a parallel-based deep fuzzy system, which fused the extracted features from fuzzy logic and DNNs for further classification tasks [31]. Similarly, Pham et al. fused the extracted feature from fuzzy logic and multiple DNNs for modeling [32].

While both types of deep ensemble TSK-FS models have demonstrated great performance, they are confronted with critical challenges. Firstly, the first type of method requires the construction of several TSK-FS models to effectively handle complex data scenarios. This poses a challenge in maintaining optimal model performance while ensuring a concise model structure. Secondly, the second type of method faces the drawback of reduced transparency and interpretability in the model optimization process, stemming from the black-box nature of DNNs directly integrated into the model. Thirdly, the majority of existing deep TSK-FS models are primarily designed for classification tasks rather than being universally applicable across diverse application scenarios. The new differentiable TSK-FS for representation learning proposed in this paper will address these issues.

*C. Iterative Shrinkage-thresholding Differentiable Optimization*

In machine learning, it is a common to enforce a sparse constraint by introducing the lasso constraint to the objective function. Consider linear regression, which can be defined as:

$$\min_{\mathbf{W}} \|\mathbf{XW} - \mathbf{Y}\|_2 + \alpha \|\mathbf{W}\|_1 \tag{4}$$

where $\mathbf{X} \in R^{N \times d}$ is the input data, $\mathbf{W} \in R^{d \times c}$ is the mapping matrix, and $\mathbf{Y} \in R^{N \times c}$ is the label matrix. Here, *N* is the number of instances, *d* is the feature dimension of input data, *c* is the number of classes, and $\alpha$ is the regularization parameter. There are two terms in (4), the first term corresponds to linear regression empirical error and the second term represents the lasso constraint. However, the lasso constraint is locally non-differentiable. To address this issue, several effective differentiable optimization methods have been proposed. For example, Daubechies et al. designed an iterative shrinkage and thresholding algorithm (ISTA) [33]. Moreover, Beck et al. proposed fast ISTA [34] on the basis of ISTA, which preserves the concise structure while reducing the convergence time. In addition, Gregor et al. proposed LISTA [35], an extension of ISTA that introduces learnable parameters into the optimization process to further improve performance.

Taking ISTA as an example, (4) can be solved by the following iterative updating rules:

$$\mathbf{W}_k = \mathbf{Pro}_{\frac{\alpha}{L}}\left(\mathbf{W}_{k-1} - \frac{1}{L}\nabla \mathcal{L}(\mathbf{W}_{k-1})\right) \tag{5}$$

$$\left[\mathbf{Pro}_{\frac{\alpha}{L}}(\mathbf{V})\right]_i = \begin{cases} \mathbf{V}_i - \alpha/L, & if\ \mathbf{V}_i > \alpha/L \\ 0, & if\ |\mathbf{V}_i| \leq \alpha/L \\ \mathbf{V}_i + \alpha/L, & if\ \mathbf{V}_i < -\alpha/L \end{cases} \tag{6}$$

where $k=1,2, ..., K$, *K* is the number of iterations. $\nabla \mathcal{L}(\mathbf{W}_{k-1}) = \mathbf{X}^T(\mathbf{XW}_{k-1} - \mathbf{Y})$, $\mathbf{Pro}_{\frac{\alpha}{L}}(*)$ is the proximal operator, $\mathbf{V} = \mathbf{W}_{k-1} - \frac{1}{L}\nabla \mathcal{L}(\mathbf{W}_{k-1})$, *L* equal to the upper limit of the maximum eigenvalue of $\mathbf{X}^T\mathbf{X}$, and $\alpha/L$ is the soft threshold. Then, (5) can be transformed as follows:

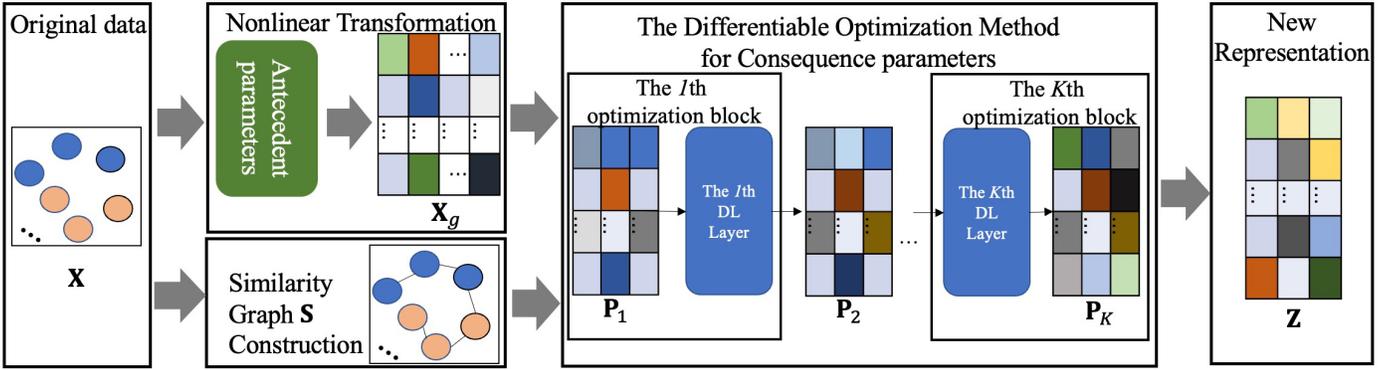

Fig. 1 The framework of the proposed method. **X** is the input data, **S** is the similarity graph, $\mathbf{X}_g$ is the representation in fuzzy feature space, $\mathbf{P}_k$ is the consequent parameters of the $k$th optimization block, and **Z** is the optimal low dimensional representation.

$$\mathbf{W}_k = \mathbf{Pro}_{\frac{\alpha}{L}}\left(\mathbf{W}_{k-1} - \frac{1}{L}\mathbf{X}^T(\mathbf{X}\mathbf{W}_{k-1} - \mathbf{Y})\right)$$
$$= \mathbf{Pro}_{\frac{\alpha}{L}}\left(\left(\mathbf{I} - \frac{1}{L}\mathbf{X}^T\mathbf{X}\right)\mathbf{W}_{k-1} - \frac{1}{L}\mathbf{X}^T\mathbf{Y}\right) \quad (7)$$

By iteratively solving the optimization (7) until convergence, a local optimal solution can be obtained. In this paper, this optimization process is followed and improved for learning the consequence parameters of TSK-FS. Then, a new rule-based differentiable representation learning method is proposed, providing a fresh perspective on integrating TSK-FS and DNN.

### III. FUZZY RULE-BASED DIFFERENTIABLE REPRESENTATION LEARNING

#### A. The Proposed Method Framework

In this section, a new fuzzy rule-based representation learning method (FRDRL) is proposed to tackle the challenges discussed above. The framework of FRDRL is illustrated in Fig. 1. As depicted in Fig. 1, the FRDRL first assesses the antecedent parameters and constructs the similarity graph from the original data. Subsequently, the original data is mapped into a high-dimension fuzzy space, generating a new fuzzy representation. The next step involves the integration of the fuzzy representation and the similarity graph into the proposed differentiable optimization method to obtain the optimal low-dimensional representation. The differentiable optimization method consists of several optimization blocks, with each block incorporating a differentiable learning layer dedicated to the learning of consequent parameters.

#### B. Representation Learning with TSK Fuzzy System

In non-linear transformation-based representation learning, the typical procedure involves mapping data into a high-dimension feature space and subsequently deriving a concise, yet highly discriminative, low dimensional representation. Interestingly, this workflow can also be realized within the framework of TSK-FS using the IF-THEN rule structure. More specifically, the IF part of TSK-FS can effectively map the original data into a high-dimensional fuzzy feature space, while the THEN part of TSK-FS can be used to extract the desired low dimensional representation.

To construct a TSK-FS for representation learning, the first step involves constructing the IF part of TSK-FS, which estimates the antecedent parameters from the original data and subsequently map the data into a fuzzy feature space. A crucial step in this process is the selection of the appropriate membership function. Previous research indicates that the Gaussian function is a reliable and commonly adopted choice [16]. Consequently, in FRDRL, the Gaussian function is used as a membership function, with the antecedent parameters representing the center and width of the Gaussian function, as shown in (2a). The center parameters of the Gaussian function $\mathbf{E} = [e_{h,j}]_{h \times d}$ can be estimated using a clustering algorithm, such as FCM, once the number of fuzzy rules $H$ is determined. Here, $d$ is the feature dimension. After it, the width parameters of the Gaussian function $\mathbf{Q} = [q_{h,j}]_{h \times d}$ is estimated as follows:

$$q_{h,j} = \sum_{i=1}^{N}(x_{i,j} - e_{h,j})^2 / \sum_{h'=1}^{H}\sum_{i=1}^{N}(x_{i,j} - e_{h',j})^2 \quad (8)$$

where $j = 1, 2, \ldots, d$, $x_{i,j}$ is the $j$th feature of $i$th instance.

Given an instance $\mathbf{x} \in R^{1 \times d}$, it can be transformed into fuzzy feature space as (9a) by using (3b) - (3d) when the center and width parameters of the TSK-FS have been estimated.

$$\mathbf{x}_g = [\tilde{\mathbf{x}}^1, \tilde{\mathbf{x}}^2, \ldots, \tilde{\mathbf{x}}^H] \in R^{1 \times H(d+1)} \quad (9a)$$

When the data are transformed into fuzzy feature space, the THEN part of TSK-FS, i.e., the consequent parameters, can be constructed for low dimensional representation extracting. Denoting traditional multi-output TSK-FS as feature transformation $\phi(*)$, the following function can be obtained:

$$\phi(\mathbf{x}) = \mathbf{x}_g \mathbf{P} \quad (9b)$$

$$\mathbf{P} = [\mathbf{p}_g^1, \mathbf{p}_g^2, \ldots, \mathbf{p}_g^m] \in R^{H(d+1) \times m} \quad (9c)$$

where $m$ is the feature dimension after transform. Different from (3a), (9c) consists of multiple group consequent parameters in each fuzzy rule. In this way, the new low dimensional representation of a given data $\mathbf{X} \in R^{N \times d}$, can be obtained as follows:

$$\mathbf{Z} = \mathbf{X}_g \mathbf{P} \in R^{N \times m} \quad (10)$$

$$\mathbf{X}_g = [\mathbf{x}_{g,1}; \mathbf{x}_{g,2}; \ldots; \mathbf{x}_{g,N}] \in R^{N \times H(d+1)} \quad (11)$$

where $\mathbf{X}_g$ is the representation transformed by fuzzy rule, and **Z** is the extracted low dimensional representation.

To further improve the performance of representation learning, the graph-based method is commonly used to preserve the geometric structure from the original feature space [36], and its objective function is given as follows:

$$\min_{\mathbf{P}} \sum_{i=1}^{N}\sum_{j=1}^{N} s_{i,j}\|\mathbf{z}_i - \mathbf{z}_j\|_2 = tr(\mathbf{Z}^\mathrm{T}\mathbf{L}\mathbf{Z}) \quad (12)$$

where $\mathbf{S} = [s_{i,j}]_{N\times N}$ is the similarity matrix which is estimated in the original feature space and $s_{i,j}$ is estimated as follows:

$$s_{i,j} = \begin{cases} \kappa(\mathbf{x}_{g,i}, \mathbf{x}_{g,j}), & \text{if } i\text{-th instance is the } k\text{-nearest} \\ & \text{neighbor of } j\text{-th instance} \\ 0, & \text{otherwise} \end{cases} \quad (13)$$

where $\kappa(*,*)$ is a kernel function, $\mathbf{L} = \mathbf{D} - \mathbf{S} \in R^{N\times N}$ is the Laplacian matrix, and $\mathbf{D} \in R^{N\times N}$ is the diagonal matrix with the $i$th diagonal value $d_{i,i} = \sum_{j=1}^{N} s_{i,j}$.

Obviously, (12) only examines the similarity information between two instances, which might not be adequate. In order to enhance the preservation of the geometric structure quality and the discriminability of the low-dimensional representation, it is essential to explore the similarity information between an instance and its neighboring instance sets. To achieve this objective, this paper introduces a second-order graph-based method for preserving geometric structure. The objective function of this method is formulated as follows:

$$\min_{\mathbf{P}} \sum_{i=1}^{N}\sum_{j=1}^{N} s_{i,j}\|\mathbf{z}_i - \mathbf{z}_j\|_2 + \sum_{i=1}^{N}\|\mathbf{z}_i - \sum_{j=1}^{N} s_{i,j}\mathbf{z}_j\|_2 = tr(\mathbf{Z}^\mathrm{T}\mathbf{L}\mathbf{Z}) + tr(\mathbf{Z}^\mathrm{T}\mathbf{\Lambda}^\mathrm{T}\mathbf{\Lambda}\mathbf{Z}) \quad (14)$$

where $\mathbf{\Lambda} = \mathbf{I} - \mathbf{S} \in R^{N\times N}$, $\mathbf{I} \in R^{N\times N}$ is the identity matrix. Finally, a sparsity constraint is further introduced to remove the redundant information and the updated objective function as follows:

$$\min_{\mathbf{P}} tr(\mathbf{Z}^\mathrm{T}\mathbf{L}\mathbf{Z}) + tr(\mathbf{Z}^\mathrm{T}\mathbf{\Lambda}^\mathrm{T}\mathbf{\Lambda}\mathbf{Z}) + \alpha\|\mathbf{P}\|_1 \quad (15)$$

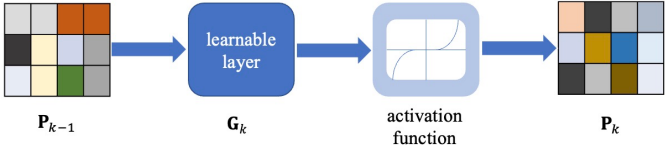

Fig. 2. One differentiable optimization block.

## C. Differentiable Optimization Method

In this subsection, the optimization process of (15) is described in detail. Since the optimization problem in (15) is locally non-differentiable, (15) can be expressed as follows by ignoring the sparsity constraint:

$$\mathcal{L}(\mathbf{P}) = \min_{\mathbf{P}} tr(\mathbf{Z}^\mathrm{T}\mathbf{L}\mathbf{Z}) + tr(\mathbf{Z}^\mathrm{T}\mathbf{\Lambda}^\mathrm{T}\mathbf{\Lambda}\mathbf{Z}) \quad (16)$$

Taking the derivative with respect to $\mathbf{P}$, we get:

$$\frac{\partial \mathcal{L}}{\partial \mathbf{P}} = \left(\mathbf{X}_g^\mathrm{T}\mathbf{L}\mathbf{X}_g + \mathbf{X}_g^\mathrm{T}\mathbf{\Lambda}^\mathrm{T}\mathbf{\Lambda}\mathbf{X}_g\right)\mathbf{P} \quad (17)$$

Next, based on the analysis in subsection II-C, the iterative shrinkage and thresholding algorithm can be applied to non-smooth $L_1$-regularized problems, and the process of calculating $\mathbf{P}$ is written as:

$$\mathbf{P}_k = \mathbf{Pro}_{\frac{\alpha}{L}}\left(\mathbf{P}_{k-1} - \frac{1}{L}\left(\mathbf{X}_g^\mathrm{T}(\mathbf{L} + \mathbf{\Lambda}^\mathrm{T}\mathbf{\Lambda})\mathbf{X}_g\right)\mathbf{P}_{k-1}\right) =$$

$$\mathbf{Pro}_{\frac{\alpha}{L}}\left(\left(\mathbf{I} - \frac{1}{L}\left(\mathbf{X}_g^\mathrm{T}(\mathbf{L} + \mathbf{\Lambda}^\mathrm{T}\mathbf{\Lambda})\mathbf{X}_g\right)\right)\mathbf{P}_{k-1}\right) \quad (18)$$

where $k$ represents the $k$th optimization iteration.

While the solution of (15) can be obtained by iteratively solving (18), the limited capacity to learn nonlinear consequence parameters restricts the complex data mining capability. Moreover, the space for computing the solution is constrained due to most of the parameters in (18) are fixed. It is an interesting question of how to expand the solution space of (18) and how to enhance the nonlinear data mining ability of the consequence parameters to further improve the performance. In [37], Lu et al. expand the solution space by linking traditional machine learning with deep learning. Specifically, they replaced some of the parameters in the traditional optimization process with learnable layers and transformed non-negative constraints into the RELU activation function. Inspired by [37], a similar attempt is conducted in this paper. Fortunately, when $\left(\mathbf{I} - \frac{1}{L}\left(\mathbf{X}_g^\mathrm{T}(\mathbf{L} + \mathbf{\Lambda}^\mathrm{T}\mathbf{\Lambda})\mathbf{X}_g\right)\right)$ is replaced as a learnable layer $\mathbf{G}_k$ and $\frac{\alpha}{L}$ is as replaced a learnable variable $\theta$. (18) equals the following:

$$\mathbf{P}_k = \mathbf{Pro}_\theta(\mathbf{G}_k\mathbf{P}_{k-1}) \quad (19)$$

Similar to [37], iteration (19) is quite similar to a block of DNN. Meanwhile, according to (6) in subsection II-C, $\mathbf{Pro}_\theta(*)$ can be transformed into the consisting of some non-linear activation functions (e.g. RELU [38]).

Building upon the aforementioned analysis, the specifics of a differentiable learning block are shown in Fig. 2. As depicted in Fig. 2, $\mathbf{P}_{k-1}$ is sequentially updated by a learnable layer and an activation function to obtain the updated $\mathbf{P}_k$. Subsequently, the optimal consequent parameters $\mathbf{P}_K$ can be obtained after $K$ optimization layers. In this way, the TSK-FS and DNN are linked in a new approach, which ensures a concise and transparent model in comparison to existing deep TSK-FS models. Moreover, the learning-based optimization adds perturbations to the traditional optimization process, expanding the search space of solutions, and thus yielding improved solutions.

## D. Loss Function for Downstream Task

In this paper, a representation learning method is proposed. Therefore, the loss function can be determined according to different downstream tasks. For example, denoting the last layer of output $\mathbf{P}_K$ as $\mathbf{P}$, the loss function can be defined as follows when the downstream task is classification:

$$L_{classification} = \|softmax(\mathbf{Z}) - \mathbf{Y}\|_2^2 + \beta\|\mathbf{P}\|_2 \quad (20)$$

where $\mathbf{Z} = \mathbf{X}_g\mathbf{P}$, $\mathbf{Y} \in R^{N\times c}$ is the label matrix, $c$ is the number of classes, and $\beta$ is the regularization parameter, it is set as 0.01 in this paper. This is a possible solution, and it is applicable when replaced by other classification loss functions.

For another example, the loss function can be defined as follows when the downstream task is clustering:

$$L_{clustering} = \|\mathbf{Z}^\mathrm{T} - \mathbf{V}\mathbf{U}\|_F^2$$
$$s.t.\ \mathbf{U}_{i,j} \in \{0,1\}, \sum_{i=1}^{c}\mathbf{U}_{i,j} = 1 \quad (21)$$

where (21) is a traditional K-means algorithm [39], $\mathbf{V} \in R^{m \times C}$ is the clustering center matrix, $\mathbf{U} \in R^{C \times N}$ is the clustering partition matrix. The updated rules of the clustering center and partition matrix can be obtained as follows:

$$\mathbf{V} = (\mathbf{UZ})^{\mathrm{T}}(\mathbf{UU}^{\mathrm{T}})^{-1} \tag{22}$$

$$\mathbf{U}_{i,j} = \begin{cases} 1, i = \underset{c}{\mathrm{argmin}} \left\| \mathbf{Z}_{j,:}^{\mathrm{T}} - \mathbf{V}_{:,c} \right\|_F^2 \\ 0, \quad otherwise \end{cases} \tag{23}$$

Then, when the loss function is determined, all learnable variables ($\mathbf{G}$ and $\theta$) can be optimized by back-propagation.

Based on the above analyses, the algorithm of the proposed FRDRL for classification and clustering task is given in Algorithm 1 and Algorithm 2, respectively.

In Algorithms 1 and 2, the consequent parameters are initialized with random, the weight of each learnable layer is initialized as $\left(\mathbf{I} - \frac{1}{L}\left(\mathbf{X}_g^{\mathrm{T}}(\mathbf{L} + \mathbf{\Lambda}^{\mathrm{T}}\mathbf{\Lambda})\mathbf{X}_g\right)\right)$, $\theta$ is initialized as $\frac{\alpha}{L}$, $L$ can be estimated according to [33] and $\alpha$ is set as 0.0001.

---

**Algorithm 1 FRDRL (classification)**

**Input**: data $\mathbf{X} \in R^{N \times d}$, the number of fuzzy rules $H$, number of maximum epochs $T$, the number of block number $K$.
**Output**: the learned $\mathbf{P}_K$.
1: Initialize $\mathbf{L}, \mathbf{\Lambda}$ based on (13).
2: Use the ESSC clustering algorithm [40] to estimate the antecedent parameters of the TSK fuzzy systems based on data $\mathbf{X}$.
3: Use (3.b) - (3.d) to map the data into the fuzzy feature space $\mathbf{X}_g$.
4: Initialize $\mathbf{P}$, and learnable variables $\theta$ and $\mathbf{G}_k$, $k=1,2…, K$.
5: for $t=1, 2, …, T$ do
6:   for $k=1, 2, …, K$ do
7:     Update $\mathbf{P}_k$ based on (19).
8:   end
9:   Compute loss based on (20).
10:  Update $\mathbf{G}_k$ and $\theta$ with back propagation, $k=1,2…, K$.
11: end

---

**Algorithm 2 FRDRL (clustering)**

**Input**: data $\mathbf{X} \in R^{N \times d}$, the number of fuzzy rules $H$, number of maximum epochs $T$, the number of block number $K$.
**Output**: the learned $\mathbf{P}_K$, clustering partition matrix $\mathbf{U}$ and clustering center matrix $\mathbf{V}$.
1: Initialize $\mathbf{L}, \mathbf{\Lambda}$ based on (13).
2: Use the ESSC clustering algorithm [40] to estimate the antecedent parameters of the TSK fuzzy systems based on data $\mathbf{X}$.
3: Use (3.b) - (3.d) to map the data into the fuzzy feature space $\mathbf{X}_g$.
4: Initialize $\mathbf{P}$, and learnable variables $\theta$ and $\mathbf{G}_k$, $k=1,2…, K$.
5: for $t=1, 2, …, T$ do
6:   for $k=1, 2, …, K$ do
7:     Update $\mathbf{P}_k$ based on (19).
8:   end
9:   Update $\mathbf{V}$ based on (22).
10:  Update $\mathbf{U}$ based on (23).
11:  Compute loss based on (21).
12:  Update $\mathbf{G}_k$ and $\theta$ with back propagation, $k=1,2…, K$.
13: end

---

### E. Discussion

In recent years, DNN-based models have achieved promising performance in various applications, yet their interpretability remain a significant challenge. Therefore, effectively balancing performance and interpretability is an ongoing task. This paper addresses this challenge by establishing a connection between the rule-based model and DNN using the differentiable optimization method. Firstly, as the proposed FRDRL is constructed based on the rule-based TSK-FS model whose interpretability has been proven in previous research [41-43], the proposed FRDRL inherently possesses interpretability. Secondly, a traditional optimization algorithm is used to construct the network structure in the optimization process, ensuring the transparency of the proposed FRDRL. Simultaneously, learning-based optimization is implemented to expand the search space of solutions. These strategy enables the proposed FRDRL to strike a balance between performance and interpretability.

Next, we discuss the computational complexity of the proposed FRDRL. Taking Algorithm 1 as an example, the major computational complexities lie in step 7 and step 9. Their time complexities are $O(d_g^2 mKT)$ and $O((d_g + m)d_g NT)$, respectively, where $d_g$ is the feature dimension in fuzzy feature space, $N$ is the number of instances, $m$ is the output feature dimension of FRDRL, $K$ is the number of optimization block and $T$ is the number of epochs. Therefore, the computational complexity of the overall algorithm is $O(d_g^2 NT)$.

Table I Statistics of datasets

| Dataset | Size | Number of feature dimensions | Number of Classes |
|---|---|---|---|
| Wine | 178 | 13 | 3 |
| IS | 2310 | 19 | 7 |
| Water | 527 | 38 | 15 |
| NATICU | 29332 | 86 | 2 |
| PAG | 5473 | 10 | 5 |
| Yale | 165 | 1024 | 15 |
| ALOI | 10800 | 125 | 100 |
| EEG | 500 | 6 | 2 |

## IV. EXPERIMENTAL STUDIES

We conduct experiments to evaluate the proposed model and answer the following questions: (1) what is the classification and clustering performance of the proposed FRDRL? (2) How effective is the proposed differentiable optimization method and second-order geometric structure preservation? (3) How is the convergence of the proposed FRDRL? (4) What is the interpretability of the proposed FRDRL?

### A. Evaluation of Classification Task

#### 1) Experimental setting

**Datasets**: To comprehensively evaluate the performance of the proposed method on classification tasks, five UCI datasets, two popular image datasets (Yale[1] and ALOI[2]), and one EEG dataset [44] were adopted for experiments. A brief description of these datasets is provided below, with detailed statistics available in Table I.

- Wine dataset: This dataset uses chemical analysis to determine the origin of wines.

---

[1] http://cvc.cs.yale.edu/cvc/projects/yalefaces/yalefaces.html
[2] https:// aloi.science.uva.nl/

Table II. ACC score (mean ± SD) of comparing models on eight datasets

| Algorithms | Wine | IS | Water | NATICU | PAG | Yale | ALOI | EEG |
|---|---|---|---|---|---|---|---|---|
| PCA | 0.9830 ±0.0255 | 0.8571 ±0.0118 | 0.5882 ±0.0443 | 0.8984 ±0.0028 | 0.9443 ±0.0069 | 0.7030 ±0.0657 | 0.7406 ±0.0108 | 0.9300 ±0.0187 |
| KPCA | 0.9719 ±0.0282 | 0.8853 ±0.0118 | 0.6166 ±0.0426 | 0.8248 ±0.0044 | 0.9468 ±0.0094 | 0.7152 ±0.0460 | 0.7832 ±0.0120 | 0.9280 ±0.0084 |
| DNN | 0.9889 ±0.0248 | 0.7918 ±0.0312 | 0.6226 ±0.0743 | 0.9606 ±0.0055 | 0.9147 ±0.0167 | 0.8242 ±0.0174 | 0.7085 ±0.0603 | 0.9360 ±0.0270 |
| DAE | 0.9832 ±0.0154 | 0.8208 ±0.0112 | 0.6186 ±0.0313 | 0.9450 ±0.0040 | 0.9424 ±0.0053 | 0.7212 ±0.1162 | 0.8270 ±0.0041 | 0.9360 ±0.0351 |
| SVM | 0.9886 ±0.0256 | 0.8593 ±0.0014 | 0.4022 ±0.0481 | 0.9546 ±0.0030 | 0.9287 ±0.0014 | 0.4727 ±0.0166 | 0.2134 ±0.0178 | 0.8780 ±0.0130 |
| TSK_L2 | **0.9887** ±**0.0154** | 0.8931 ±0.0168 | 0.6032 ±0.0564 | 0.9391 ±0.0109 | 0.9547 ±0.0025 | 0.4242 ±0.0678 | 0.8676 ±0.0066 | 0.9400 ±0.0235 |
| TSK_BN_UR | 0.9222 ±0.0774 | 0.9355 ±0.0099 | 0.6189 ±0.0284 | 0.9653 ±0.0016 | 0.9523 ±0.0053 | 0.4061 ±0.0891 | 0.7951 ±0.0170 | 0.9339 ±0.0219 |
| RRL | 0.9721 ±0.0278 | 0.8824 ±0.0177 | 0.6091 ±0.0413 | 0.9563 ±0.0084 | 0.9526 ±0.0022 | 0.8242 ±0.0944 | 0.7894 ±0.0104 | 0.9140 ±0.0219 |
| FFDN | 0.9777 ±0.0236 | 0.9615 ±0.0079 | **0.6263** ±**0.0430** | 0.9614 ±0.0048 | 0.9686 ±0.0037 | 0.8061 ±0.1019 | **0.9252** ±**0.0076** | 0.9460 ±0.0167 |
| D-TSK-FC | 0.9636 ±0.0272 | 0.8797 ±0.0022 | 0.4984 ±0.0523 | 0.5470 ±0.0693 | 0.9516 ±0.0054 | 0.8000 ±0.1870 | 0.8451 ±0.0241 | 0.9376 ±0.0255 |
| FRDRL | 0.9877 ±0.0169 | **0.9623** ±**0.0061** | 0.6112 ±0.0462 | **0.9622** ±**0.0037** | **0.9720** ±**0.0025** | **0.8364** ±**0.0618** | 0.9142 ±0.0068 | **0.9580** ±**0.0160** |

Table III. mF1 score (mean ± SD) of comparing models on eight datasets

| Algorithms | Wine | IS | Water | NATICU | PAG | Yale | ALOI | EEG |
|---|---|---|---|---|---|---|---|---|
| PCA | 0.9727 ±0.0242 | 0.8126 ±0.0196 | 0.3591 ±0.0542 | 0.8959 ±0.0013 | 0.6368 ±0.0117 | 0.5535 ±0.0349 | 0.7320 ±0.0130 | 0.9276 ±0.0158 |
| KPCA | 0.9609 ±0.0428 | 0.8453 ±0.0241 | 0.4056 ±0.0676 | 0.7868 ±0.0596 | 0.6656 ±0.0393 | 0.5477 ±0.0437 | 0.7797 ±0.0120 | 0.9258 ±0.0073 |
| DNN | 0.9897 ±0.0231 | 0.7886 ±0.0324 | 0.3404 ±0.0629 | 0.9606 ±0.0055 | 0.8468 ±0.0564 | 0.7985 ±0.0178 | 0.6457 ±0.0639 | 0.9337 ±0.0276 |
| DAE | 0.9824 ±0.0161 | 0.8177 ±0.0110 | 0.3677 ±0.0420 | 0.9450 ±0.0040 | 0.6226 ±0.0255 | 0.7192 ±0.1215 | 0.8315 ±0.0039 | 0.9331 ±0.0361 |
| SVM | 0.9876 ±0.0226 | 0.8791 ±0.0145 | 0.1312 ±0.0467 | 0.9550 ±0.0025 | 0.3061 ±0.0243 | 0.4537 ±0.0284 | 0.1551 ±0.0117 | 0.8395 ±0.0181 |
| TSK_L2 | 0.9507 ±0.0589 | 0.8917 ±0.0189 | 0.3572 ±0.0573 | 0.9368 ±0.0093 | 0.6808 ±0.0272 | 0.4363 ±0.0674 | 0.7954 ±0.0148 | 0.9378 ±0.0215 |
| TSK_BN_UR | 0.9241 ±0.0757 | 0.9341 ±0.0117 | 0.3775 ±0.0623 | 0.9653 ±0.0022 | 0.6584 ±0.0476 | 0.3634 ±0.0898 | 0.7901 ±0.0109 | 0.9305 ±0.0196 |
| RRL | 0.9729 ±0.0269 | 0.8820 ±0.0169 | 0.3511 ±0.0556 | 0.9563 ±0.0084 | 0.7200 ±0.0154 | 0.7754 ±0.1031 | 0.7846 ±0.0087 | 0.9100 ±0.0246 |
| FFDN | 0.9776 ±0.0325 | 0.9615 ±0.0080 | 0.3998 ±0.0761 | 0.9615 ±0.0048 | 0.7630 ±0.0450 | 0.8004 ±0.1065 | **0.9191** ±**0.0111** | 0.9324 ±0.0207 |
| D-TSK-FC | 0.9641 ±0.0281 | 0.8783 ±0.0261 | 0.2159 ±0.0813 | 0.4328 ±0.1267 | 0.6926 ±0.0633 | 0.5428 ±0.2285 | 0.4910 ±0.0701 | 0.9355 ±0.0272 |
| FRDRL | **0.9884** ±**0.0159** | **0.9629** ±**0.0061** | **0.4629** ±**0.0122** | **0.9623** ±**0.0037** | **0.8517** ±**0.0151** | **0.8101** ±**0.0625** | 0.9015 ±0.0067 | **0.9650** ±**0.0135** |

- Image Segmentation (IS) dataset: this is an outdoor image dataset and consists of shape and RBG features.
- Water dataset: This dataset comes from the daily measures of sensors in an urban wastewater treatment plant.
- NATICU dataset: This is an Android malware detection dataset, consisting of benign and malware apps released between the years 2010-2019.
- PAG dataset: This is a document page layout classification dataset.
- Yale dataset: This is a face dataset consisting of 15 people's different facial expressions or configurations, and the Grayscale features are extracted for experiments.
- ALOI dataset: This is a color image collection of one-thousand small objects and the RGB feature is extracted as the feature for experiments.
- EEG dataset: This is the real epileptic EEG dataset, and the wavelet packet decomposition (WPD) is extracted as the feature for experiments.

**Methods for comparison**: Ten state-of-the-art methods were compared with the proposed method in the classification experiments, including two classical unsupervised representation learning methods (PCA [4], KPCA [8]), two deep unsupervised representation learning methods (DNN [45], DAE [7]), both constructed based on Multilayer Perceptron (MLP), three interpretable traditional classification methods (SVM [46], TSK_L2 [47], TSK_BN_UR [48]), and three deep rule-based classification methods (RRL [49], FFDN [31] and D-TSK-FC [24]). 5-fold cross-validation was adopted to evaluate the classification performance more fairly, and ACC and mean F1 score (mF1) [50] were used as evaluation metrics to evaluate the classification performance.

**Implementation details of classification task**: In the proposed FRDRL, the number of rules is varied from 2 to 20,

Table IV. NMI score (mean ± SD) of comparing models on five datasets

| Algorithms | IS | Water | NATICU | PAG | ALOI |
|---|---|---|---|---|---|
| PCA | 0.6100±0.0026 | 0.1311±0.0137 | 0.2827±0.0000 | 0.0859±0.0000 | 0.6187±0.0091 |
| KPCA | 0.5806±0.0000 | 0.1417±0.0072 | 0.2016±0.0000 | 0.0815±0.0000 | 0.6377±0.0087 |
| DNN | 0.4621±0.0349 | 0.1588±0.0278 | 0.3427±0.0049 | 0.1739±0.0166 | 0.6529±0.0062 |
| DAE | 0.4281±0.0308 | 0.1449±0.0123 | 0.2676±0.0633 | 0.1064±0.0124 | 0.6479±0.0135 |
| K-means | 0.5045±0.0004 | 0.1661±0.0056 | 0.2163±0.0000 | 0.0881±0.0005 | 0.6293±0.0087 |
| EE-K-means | 0.4872±0.0185 | 0.1639±0.0000 | 0.2666±0.0000 | 0.0872±0.0000 | 0.6292±0.0050 |
| DFKM | 0.5164±0.0298 | **0.1876±0.0196** | 0.2175±0.0617 | 0.1448±0.0032 | 0.6619±0.0173 |
| FRDRL | **0.6381±0.0112** | 0.1802±0.0076 | **0.3615±0.0063** | **0.2300±0.0049** | **0.6939±0.0011** |

Table V. ARI score (mean ± SD) of comparing models on five datasets

| Algorithms | IS | Water | NATICU | PAG | ALOI |
|---|---|---|---|---|---|
| PCA | 0.5029±0.0025 | 0.0733±0.0072 | 0.3653±0.0000 | 0.0877±0.0000 | 0.1648±0.0125 |
| KPCA | 0.4717±0.0000 | 0.0739±0.0043 | 0.2575±0.0000 | 0.0890±0.0000 | 0.2190±0.0347 |
| DNN | 0.4621±0.0349 | 0.0647±0.0172 | 0.4205±0.0264 | 0.1064±0.0109 | 0.1893±0.0002 |
| DAE | 0.4281±0.0308 | 0.0573±0.0046 | 0.3413±0.0793 | 0.0662±0.0234 | 0.1969±0.0283 |
| K-means | 0.5045±0.0004 | 0.1070±0.0117 | 0.2839±0.0000 | 0.1080±0.0006 | 0.1847±0.0116 |
| EE-K-means | 0.4872±0.0185 | 0.1073±0.0000 | 0.3462±0.0000 | 0.1075±0.0000 | 0.1725±0.016 |
| DFKM | 0.5165±0.0298 | 0.0851±0.0202 | 0.2746±0.0701 | 0.1104±0.0184 | **0.2998±0.0425** |
| FRDRL | **0.5286±0.0086** | **0.1113±0.0036** | **0.4575±0.0091** | **0.2025±0.0001** | 0.2521±0.0012 |

with a step size of 2. The number of optimization blocks is set in the range of 5 to 20 with a step size of 5. The output feature dimension of each block, denoted as *m*, is set as the number of classes. The MLP is used as each learnable layer, with the learning rate chosen from the range of [1e-5, 1e-4], and the number of epochs selected from [100,500,1000,1500]. A grid search strategy is used to determine the optimal parameters in this experiment. Detailed parameter settings for the comparison algorithms can be found in the *Supplementary Materials* section.

*2) Classification Performance*

The experimental results are shown in Tables II-III. From the results, the following observations can be obtained.
(1) Since FRDRL introduces the second-order geometric structure preservation and proposes a traditional optimization algorithm to construct the network structure, which realizes the learning-based optimization, this leads FRDRL to obtain better performance compared to existing methods while ensuring model interpretability.

(2) Compared with four representation learning methods, FRDRL demonstrates superior classification performance. The two traditional representation learning methods (PCA and KPCA) perform moderately well on datasets with fewer classes and feature dimensions, but they exhibit inferior performance when handling larger datasets such as the Yale and ALOI datasets. Meanwhile, while the two deep representation learning methods (DNN and DAE) achieve better performance on most datasets, they still fall short in comparison to FRDRL, primarily due to their inability to effectively explore geometric structure information.

(3) Three traditional interpretable classification algorithms can indeed achieve better performance on simpler datasets, such as the wine dataset. However, on more complex datasets, such as the Yale and ALOI datasets, these three methods fall behind FRDRL and other deep learning-based methods. This observation suggests that constructing a deep structure for modeling has positive implications for improving performance.

(4) RRL and D-TSK-FC construct their models by stacking multiple rule layers, but they generally exhibit inferior performance compared to FRDRL and FFDN across most datasets. This arises from the fact that FRDRL and FFDN integrate deep networks into their modeling. Also, FFDN neglects the geometric structure information, which can also result in its inferior performance compared to FRDRL in most datasets.

*B. Evaluation of Clustering Task*

*1) Experimental setting*

**Datasets and methods for comparison**: To further evaluate the performance of RDRL, clustering experiments were conducted on five datasets (IS, Water, NATICU, PAG, and ALOI). Four unsupervised representation learning methods (PCA, KPCA, DNN, and DAE), two traditional clustering methods (K-means [51] and EE-K-means[52]), and one deep clustering method (DFKM [53]) were adopted for comparison. The clustering performance was evaluated using the two commonly used indices NMI and ARI [54]. Moreover, each algorithm was executed 10 times with different parameters, and the best results in terms of the mean and the standard deviation of the metrics were recorded for comparison.

**Implementation details of clustering task**: In the proposed FRDRL, the number of rules was varied from 2 to 10, with a step size of 2. The number of optimization blocks was set from 1 to 9 with a step size of 2, and the output feature dimension of each block, denoted as *m*, was selected from the set [10, 20, 40, 60, 80, 100]. The MLP was used as each learnable layer, with the learning rate selected from the set [1e-5, 1e-4], and the number of epochs selected from the set [10,50,100]. A grid search strategy was used to determine the optimal parameters in this experiment. The detailed parameter settings for the comparison algorithms can be found in the *Supplementary Materials* section.

*2) Clustering Performance*

The clustering experimental results are shown in Tables IV-V. From the results, the following observations can be obtained.

Table VI. The ACC score (mean ± SD) of FRDRL, FRDRL1 and FRDRL2 on eight datasets

| Algorithms | Wine | IS | Water | NATICU | PAG | Yale | ALOI | EEG |
|---|---|---|---|---|---|---|---|---|
| FRDRL1 | 0.8205 ±0.0103 | 0.9411 ±0.0130 | 0.5294 ±0.0434 | 0.8219 ±0.0113 | 0.9506 ±0.0053 | 0.1333 ±0.0732 | 0.8804 ±00083 | 0.7660 ±0.0427 |
| FRDRL2 | 0.9832 ±0.0154 | 0.9190 ±0.0079 | 0.6111 ±0.0394 | 0.9610 ±0.0043 | 0.9693 ±0.0061 | 0.7091 ±0.0664 | **0.9119 ±0.0104** | 0.9540 ±0.0219 |
| FRDRL | **0.9877 ±0.0169** | **0.9623 ±0.0061** | **0.6112 ±0.0462** | **0.9622 ±0.0037** | **0.9720 ±0.0025** | **0.8364 ±0.0618** | 0.9142 ±0.0068 | **0.9580 ±0.0160** |

Table VII. The mF1 score (mean ± SD) of FRDRL, FRDRL1 and FRDRL2 on eight datasets

| Algorithms | Wine | IS | Water | NATICU | PAG | Yale | ALOI | EEG |
|---|---|---|---|---|---|---|---|---|
| FRDRL1 | 0.8147 ±0.0111 | 0.9408 ±0.0131 | 0.3182 ±0.0486 | 0.8191 ±0.0105 | 0.6962 ±0.0364 | 0.1079 ±0.0632 | 0.8845 ±0.0079 | 0.8216 ±0.0341 |
| FRDRL2 | 0.9827 ±0.0158 | 0.9192 ±0.0078 | 0.3618 ±0.0608 | 0.9608 ±0.0045 | **0.8459 ±0.0196** | 0.6836 ±0.0784 | **0.9026 ±0.0134** | 0.9619 ±0.0181 |
| FRDRL | **0.9884 ±0.0159** | **0.9629 ±0.0061** | **0.4629 0.0122** | **0.9623 ±0.0037** | 0.8517 ±0.0151 | **0.8101 ±0.0625** | 0.9015 ±0.0067 | **0.9650 ±0.0135** |

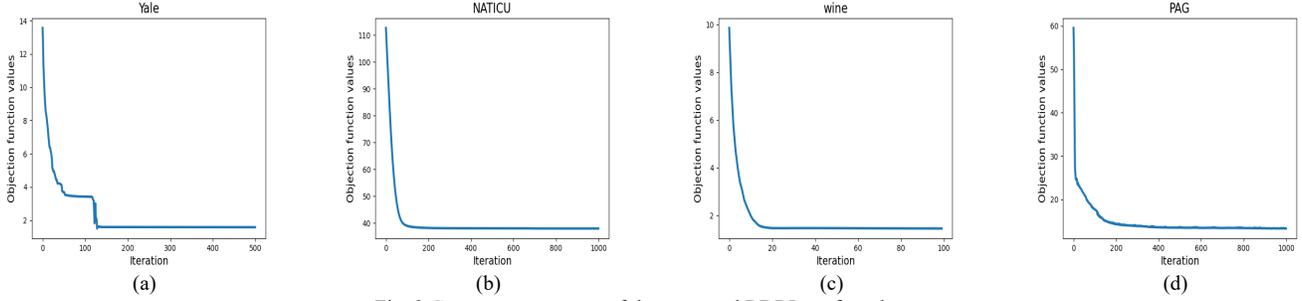

Fig. 3 Convergence curves of the proposed RDRL on four datasets.

(1) Since the proposed FRDRL is built with rule-based TSK-FS and introduces second-order geometric structure preservation, FRDRL demonstrates an exceptional capability to efficiently extract nonlinear relationships from data, leading to great performance in clustering tasks.

(2) Compared with four unsupervised representation learning methods, FRDRL exhibits superior performance in most cases. For example, on the PAG dataset, the NMI values of FRDRL are on average 10% higher than those of the four methods. This advantage can be attributed to the fact that these methods do not effectively leverage geometric structure information. In addition, the findings suggest that the proposed method possesses a more robust nonlinear mining ability.

(3) Compared with traditional clustering methods, both DFKM and FRDRL demonstrate superior performance across all datasets, highlighting the insufficiency of direct clustering on original data. Moreover, FRDRL outperforms DFKM on most datasets, indicating the effectiveness of constructing a deep model with TSK-FS for mining the nonlinear relationships between data.

### C. Ablation Studies

To further evaluate the effectiveness of the proposed differentiable optimization method and second-order geometric structure preservation, in this subsection, we define FRDRL using the traditional optimization method as FRDRL1 and define FRDRL using only the first-order geometric structure preservation as FRDRL2.

The classification experimental results are shown in Tables VI-VII. It can be seen from the results that the performance of FRDRL1 is inferior to FRDRL2 and FRDRL, indicating that the differentiable optimization method can significantly improve the performance. Furthermore, although FRDRL2 outperforms FRDRL on the PAG and ALOI datasets, it shows no significant advantages on other datasets, suggesting that the second-order geometric structure preservation has a positive influence in most cases.

### D. Convergence Analysis

In this subsection, we conducted convergence experiments on four datasets to further evaluate the convergence of FRDRL. Taking the classification task as an example, the experimental results are shown in Fig. 3. It is evident from Fig. 3 that FRDRL exhibits high convergence rates on four datasets. Specifically, on the smaller datasets (Yale and wine datasets), FRDRL can converge within 200 iterations, while on the larger datasets (NATICU and PAG), convergence is achieved within 400 iterations. Therefore, FRDRL demonstrates excellent convergence performance.

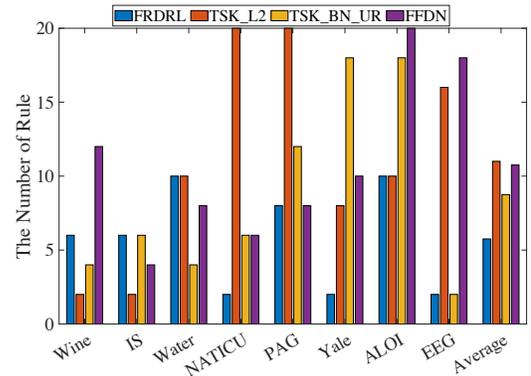

Fig 4. The rule number of four algorithms on all datasets.

## E. Interpretability Analysis

In this subsection, the interpretability of the proposed FRDRL is illustrated from two perspectives. First, the details of the modeling process translated into a natural logic language are provided for a classification task. Also, the number of rules, a commonly used measure for the interpretability of fuzzy systems, is discussed. It is generally understood that the lower the number of rules, the more interpretable the model is [17, 55]. Therefore, a comparison is conducted with several other fuzzy logic-based models in terms of the rule number.

To demonstrate the interpretability of the proposed FRDRL, we utilize the EEG dataset in the classification task as a case study. The interpretability of traditional fuzzy systems is mainly attributed to the rule-based knowledge expression and the fuzzy inference mechanism. Similarly, in this paper, the representation learning process is interpreted using a set of fuzzy rules, akin to traditional methods.

In the proposed method, the multiple outputs of the TSK fuzzy system represent the new features. The rules for feature transformation can be formulated as follows:

$$IF\ x_1\ is\ A_1^h(e_1^h, q_1^h) \wedge \cdots \wedge x_d\ is\ A_d^h(e_d^h, q_d^h)$$
$$THEN\ f^h(\mathbf{x}) = [p_0^{h,1} + p_1^{h,1}x_1 + \cdots + p_d^{h,1}x_d, p_0^{h,2} + p_1^{h,2}x_1 + \cdots + p_d^{h,2}x_d, \dots, p_0^{h,m} + p_1^{h,m}x_1 + \cdots + p_d^{h,m}x_d],\ h = 1, \dots, H \quad (24)$$

In the IF-part of the TSK fuzzy system, each fuzzy set $A_j^h(e_j^h, q_j^h)$, associated with the $j$th dimension in the $h$th rule, can be interpreted as a linguistic description based on the order of the center of the Gaussian function. For this experiment, the number of fuzzy rules is set as two, i.e., there are two clustering centers for each feature dimension of the EEG dataset. Following the existing methods [16, 41], the linguistic terms of the corresponding fuzzy sets can be represented as *Low* and *High* based on the order of the center value. In different application scenarios, the interpretation of the IF-part of TSK-FS varies, and here we provide a possible common interpretation. Subsequently, based on the antecedent parameters, i.e., $e_j^h, q_j^h$, the membership functions and the linguistic explanation of each fuzzy set are shown in Fig. S1 of the *Supplementary Materials* section. Taking the first feature of EEG as an example, it is evident from the first row in Fig. S1 that the center and variance of the first fuzzy set are (0.3733) and (0.0467), where the center value (0.3733) is larger than the second center value (0.2476). Therefore, the first fuzzy set can be expressed as *High*. Similarly, other features can be interpreted in a similar manner. By describing all fuzzy sets of all features using linguistic terms in this manner, the proposed fuzzy system for representation learning can be explained by fuzzy rules, as presented in Table VIII.

We then conducted a comparison of the number of rules for the four fuzzy system-based classification models across all datasets. As D-TSK-FC is constructed based on the zero-order fuzzy system, while the others are constructed based on the first-order fuzzy system, D-TSK-FC is excluded from the comparison. The results of the comparison are shown in Fig. 4. Fig. 4 displays the number of rules for the four algorithms across all datasets, along with the corresponding average number of rules. Analyzing Fig. 4 together with Tables II-III, it becomes apparent that the proposed algorithm, FRDRL, achieves superior results on most datasets with a reduced number of rules in comparison to other algorithms. Consequently, our algorithm demonstrates a substantial advantage in both performance and interpretability.

Table VIII. Rule base generated for EEG dataset

| The fuzzy rule-base differentiable representation learning |
|---|
| **Rule 1**: |
| **IF**: the 1th feature is High and |
|     the 2th feature is Low and |
|     the 3th feature is Low and |
|     the 4th feature is Low and |
|     the 5th feature is High and |
|     the 6th feature is High. |
| **Then**: the 1th output is $-65.811-50.558x_1+23.871x_2-12.136x_3+12.853x_4+21.709x_5-13.686x_6$ and |
|     the 2th output is $98.492+77.983x_1-35.510x_2+19.320x_3-20.164x_4-32.886x_5+20.790x_6$. |
| **Rule 2**: |
| **IF**: the 1th feature is Low and |
|     the 2th feature is High and |
|     the 3th feature is High and |
|     the 4th feature is High and |
|     the 5th feature is Low and |
|     the 6th feature is Low. |
| **Then**: the 1th output is $-44.499-4.1975x_1+36.571x_2+4.987x_3+25.764x_4+14.957x_5+16.984x_6$ and the 2th output is $65.416+2.974x_1-56x_2-5.146x_3-39.366x_4-21.545x_5-25.351x_6$. |

## V. CONCLUSION

This paper introduces a novel fuzzy rule-based representation learning method that integrates a differentiable optimization approach. This method first leverages the antecedent part of the TSK-FS to map the data into a high-dimension fuzzy space. Subsequently, it employs a differentiable optimization method for learning the consequence parameters, enabling the extraction of low-dimensional representations. In this proposed optimization method, the traditional optimization process is preserved while incorporating a learning-based optimization to further introduce perturbations, which expands the search space of solutions and strengthens the nonlinear data mining ability of the proposed method. Moreover, second-order geometric structure preservation is also introduced to improve the robustness of the method. The experimental results demonstrate that the proposed method outperforms numerous existing algorithms, highlighting its superior performance representation learning.

Despite achieving improved performance, the proposed method encounters potential challenges that warrant future research. Firstly, in geometric structure preservation, constructing the similarity matrix is time-consuming and unfriendly to large datasets, posing an interesting research problem on how to alleviate this issue. Secondly, exploring more effective mechanisms to further improve the performance of the proposed method is crucial. Lastly, extending the proposed method to other application scenarios, such as transfer learning, and multi-modal learning is worthy of further research.

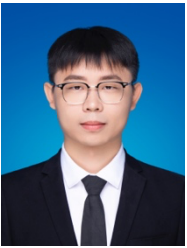

**Wei Zhang** is currently pursuing the Ph.D. degree in the School of Artificial Intelligence and Computer Science, Jiangnan University, Wuxi, China.

His research interests include computational intelligence, machine learning, interpretable artificial intelligence, fuzzy modeling and their applications.

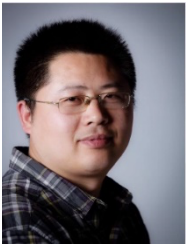

**Zhaohong Deng** (M'12-SM'14) received the B.S. degree in physics from Fuyang Normal College, Fuyang, China, in 2002, and the Ph.D. degree in information technology and engineering from Jiangnan University, Wuxi, China, in 2008.

He is currently a Professor with the School of Artificial Intelligence and Computer Science, Jiangnan University. He has visited the University of California-Davis and the Hong Kong Polytechnic University for more than two years. His current research interests include interpretable intelligence, uncertainty in artificial intelligence and their applications. He has authored or coauthored more than 100 research papers in international/national journals.

Dr. Deng has served as an Associate Editor or Guest Editor of several international Journals, such as IEEE Trans. Emerging Topics in Computational Intelligence, *Neurocomputing*, and so on.

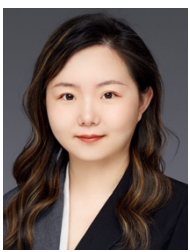

**Guanjin Wang** obtained a joint Ph.D. in computer science from the University of Technology Sydney, Australia, and in health informatics from The Hong Kong Polytechnic University, China. She is currently a senior lecturer in the School of Information Technology at Murdoch University, Perth, Australia. Her research focuses on AI, machine and deep learning, with an emphasis on handling complex and imperfect data and learning environments and achieving model explainability. She has published high-quality research papers in IEEE TFS, IEEE SMC, IEEE TCYB, and IEEE JBHI. Her research has also received recognition and support from various funding bodies, including Google Research, the governments of Australia and Hong Kong, and local state governments. Since 2021, she has served as the chapter chair in the IEEE Western Australia Section.

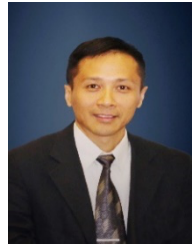

**Kup-Sze Choi** (M'97) received the Ph.D. degree in computer science and engineering from the Chinese University of Hong Kong, Hong Kong in 2004.

He is currently an Professor at the School of Nursing, Hong Kong Polytechnic University, Hong Kong, and the Director of the Centre for Smart Health. His research interests include virtual reality, artificial intelligence, and their applications in medicine and healthcare.